# Analysis of Corporate Environmental Reports using Statistical Techniques and Data Mining

Jashua Rajesh Modapothala, Swinburne University of Technology (Sarawak Campus), Kuching, Malaysia
E-mail: jmodapothala@swinburne.edu.my
Biju Issac, Swinburne University of Technology (Sarawak Campus), Kuching, Malaysia
E-mail: bissac@swinburne.edu.my

**Abstract**
*Measuring the effectiveness of corporate environmental reports, it being highly qualitative and less regulated, is often considered as a daunting task. The task becomes more complex if comparisons are to be performed. This study is undertaken to overcome the physical verification problems by implementing data mining technique. It further explores on the effectiveness by performing exploratory analysis and structural equation model to bring out the significant linkages between the selected 10 variables. Samples of 539 reports across various countries are used from an international directory to perform the statistical analysis like – One way ANOVA (Analysis of Variance), MDA (Multivariate Discriminant Analysis) and SEM (Structural Equation Modeling). The results indicate the significant differences among the various types of industries in their environmental reporting, and the exploratory factors like stakeholder, organization strategy and industrial oriented factors, proved significant. The major accomplishment is that the findings correlate with the conceptual frame work of GRI.*

## 1. Introduction

The rise of environmental reporting is observed in different forms. Traditionally, firms disclose the environmental concerns in their annual reports. The rules, procedures and presentation of these reports are too diverse across various countries [1]. Preparing such reports, in majority of the cases, is on voluntary-basis and is aimed more of a publicity than providing environmental facts and figures. The divergent information needs of the stakeholders are to be met by the reporting companies [2]. Hence, the entities have to customize the reports to different information needs and preferences. A similar view was also expressed as, 'an organization needs to send the right messages through the right distribution channels to the right audiences [3] [4]. In this context, the information is of highly qualitative in nature [5]. It is laborious and need expertise to provide more reliable information in measuring the effectiveness of environmental reports. The paper is organized as follows. Section 2 is literature review, section 3 is data mining concepts, section 4 is research objectives and hypothesis, section 5 is methodology, section 6 is results and discussion and section 7 is limitations and conclusion.

## 2. Literature Review

The environmental related information that is often presented to the external users and people-at-large via a fiscal report or some other separate report is referred as environmental reporting or corporate environmental reporting. The growing need of the information among the stakeholders and the change in organizational practices has resulted in vast amount of information which requires to be measured.

The impact of Stakeholders on companies is observed in various ways, via investments, sales, governmental pressure and ultimately the size of future business profits. Grey et al., argued that communication through these separate reports has been performed where a net benefit for the company is achievable [6]. Moreover, good environmental performance (or resource productivity) is observed as an advantage in dynamic corporate competition [7] – [9]. Therefore environmental concerns have to be integrated into corporate objectives, emphasizing that economic decisions underlie the holistic life-cycle approach to products, from the 'cradle to the grave.' 'Motives for voluntary disclosures are unlikely to be simple', and that the underlying reasons depend largely on the culture within the company. Schuster observed that environmental reporting is also an important information tool within the industrial sector in a study including Tetra Pak and Ericsson [10].

In earlier studies language used in the corporate organizations were analyzed, include the studies by Livesey and Kearins and Milne et al. [11]-[13]. A similar study in this line was conducted with all of the organization's reports since 1993 through 2003 identifying and analysing the emergence and development of a sustainable development discourse. Focusing on the language and images used to construct meanings, and the context in which the reports emerged, the traces the organization's reporting developments [14].

Studies on current voluntary corporate environmental reports meeting the requirements of the Global Reporting Initiative GRI 2000 sustainability reporting guidelines and ISO 14031 environmental performance evaluation standard were inspiration. [15]. Metrics were also used in analyzing the corporate environmental reports in view of environmental sustainability and application of taxonomies [16].

Rao et al., in their research have considered Small and Medium-size Enterprises (SMEs) operating in the food and beverage, furniture, fashion accessories, hotel and restaurant, automotive parts and electroplating sectors. The metric adopted in their research follows the framework given by the Federal Environmental Ministry in Bonn and the Federal Environmental Agency in Berlin. The empirical approach develops an exploratory analysis and a structural equation model to bring out statistically significant linkages between five latent constructs: environment management indicators, environment performance indicators, environmental performance, business performance and competitiveness [17].

In order to decide on a method for evaluating company reports, the authors looked at several potential sustainable development metrics. The frameworks reviewed included those of the World Business Council for Sustainable Development (WBCSD), the Global Environmental Management Initiative (GEMI), and the Coalition for Environmentally Responsible Economies (CERES), which convened the Global Reporting Initiative (GRI) in partnership with the United Nations Environment Programme. Eventually, have determined, GRI guidelines were the most comprehensive.

**3. Data Mining Concepts**
In coming up with the tool to evaluate these reports, data mining was chosen because of its robust in interpreting and computational. Data mining is the process of analyzing data from different sources or perspective and interpreting it into useful information. This information later can be used to increase profit from an organization's point of view, cuts costs etc. In other words, it is the process of finding correlations or relationship patterns among various fields in large relational databases.

Data mining software analyzes relationships and patterns in stored transaction data based on open-ended user queries. Generally, any of four types of relationships with data are sought [18]:

- *Classes:* Data that is stored is used to classify predetermined or predefined groups. For example, a shoe retail mart could mine customer purchase data to determine what time of the month or year customers visit and what they normally order. This information could be used to encourage customer visits by having specific type of shoes or footwear.
- *Clusters:* Data items are grouped according to logical relationships or consumer preferences. For example, the data of consumer markets can be mined to identify the current market segments or consumer preferences.
- *Associations:* Data can be mined to identify associations between events or items. For example, one grocery chain used the data mining to analyze local buying patterns. They discovered that when men bought diapers on certain days, they also tended to buy beer. The grocery chain could use this newly discovered information in various ways to increase revenue, whereby, they could move the beer display closer to the diaper display. This is an example of associative mining.
- *Sequential patterns:* Data is mined to predict behavior patterns and trends. For example, a departmental store retailer could predict the likelihood of a product being purchased based on a consumer's purchase of other related products.

Another aspect of data mining is document clustering. Here the goal is to partition documents into clusters according to their similarities. Major steps of document clustering include tokenization, stemming, elimination of stop words, index term selection etc. Tokenization divides texts into words. After tokenization, titles and abstracts are converted into words. The words from the titles are assigned twice many weights as the abstract terms. Stemming removes affixes of words. Stop words are common English words such as articles, prepositions, and conjunctions and appear in almost every article. Eliminating stop words can reduce the number of index terms and increase efficiency of document clustering algorithms. After these preprocessing steps, distinct index words that are capable of representing documents are selected.

Data mining adopted its techniques from many research areas including statistics, machine learning, association rules, neural networks etc [19]. Some are as follows:

(1) *Association Rule:* Association rule generators are an effective data mining technique used to search through an entire data set for rules to show the nature and frequency of relationships or associations between data entities. The resulting associations can be used to filter the information for human analysis and possibly to define a prediction model based on observed behavior.
(2) *Artificial Neural Networks:* They are recognized in the automatic learning framework as universal approximations, with massively parallel computing character and good generalization capabilities, but also as black boxes due to the difficulty to obtain insight into the relationship learned.
(3) *Statistical Techniques:* These include linear regression, discriminate analysis, or statistical summarization.
(4) *Machine learning (ML): I*t is the center of the data mining concept, due to its capability to gain physical insight into a problem, and participates directly in data selection and model search steps. Decision tree induction, the best known ML framework was found to be able to handle large-scale problems due to its computational efficiency, to provide interpretable results, and, in particular, able to identify the most representative attributes for a given task.

## 4. Research Objectives and Hypothesis

The overall objective of the study is to examine and analyze, in a systematic manner, the nature of environmental reporting of selected corporate companies on a global perspective. In order to perform the analysis, data mining technique is used to come up with a model to extract the data from a huge database.

1. To construct a data mining model for generating frequencies and scorecard for the annual reports
2. To examine how the selected variables have any significant impact on the industrial sector.
3. To determine which control variables discriminate between two or more naturally occurring groups.
4. To make a further investigation in exploring relationships among the factors extracted.

The industries that are subject to high level scrutiny by the environmental groups and community stake holders will have much higher environmental protection costs [20]. Thus it is hypothesized as, "The extent of reporting among the various industries is similar".

## 5. Methodology

In this study, an assessment technique designed by Haub Business and Sustainability Program (Schulich School of Business) was implemented as shown in table 1. Some modifications were made to the original technique and data mining technique was done to do the scoring for its initial measurement. The technique used in this study is similar to the most addressed issues that are required for GRI (2008) as in [21].

Table 1: Search criteria

| Search Criteria (variable) | Max. Score |
|---|---|
| Corporate environmental or HSE (health, safety and environment) policy statement. (v1) | 10 |
| Corporate policy or company views on 'sustainability' or 'sustainable development'. (v2) | 10 |
| CEO statements on 'environmental issues' and/or 'sustainable development/sustainable issues'. (v3) | 10 |
| Carbon dioxide emissions and/or global warming impact and/or climate change. (v4) | 10 |
| 'Toxic waste' and/or 'toxic emissions'. (v5) | 10 |
| 'Employee turnover' and 'employee retention' (v6) | 10 |
| Equal opportunities and/or diversity (v7) | 10 |
| References to 'human rights' (v8) | 10 |
| 'Shareholder value' and/or 'share price' and/or 'dividends' (v9) | 10 |
| Reference to 'customer satisfaction' and/or 'customer transactions' and/or 'sales' (v10) | 10 |

A study was conducted by the HBSP (Haub Business and Sustainability Program) technique in assessing online reports by tele-workers [22]. The environmental reports for the study were extracted from the Corporate Register website [23], which is a world directory of published corporate environmental and social reports. All the reports available on each of the category were initially considered. Judgmental criteria was applied to consider only those categories which got at least 30 reports (inclusive of foreign language). The time period of these reports are of 2008 only, and hence it resulted in a sample size of 690. The broad industrial sector selected for the study include: primary sector, secondary sector, and tertiary/manufacturing sector.

Table 2: Rating of scores

| Frequency (key words) | Score | Rating |
|---|---|---|
| >=75 | 10 | Very strong |
| >=50 and <75 | 7 | Strong |
| >=20 and <50 | 5 | Moderate |
| >=5 and <20 | 3 | Weak |
| <5 and >0 | 1 | Very weak |

In our study the clustering of files is done manually, but data mining is used to find groups or keyword classes. Using association rules with keywords and using statistical techniques, analysis is further done. The data mining was implemented through Java programs on over a 1000 corporate reports. It used two approaches, yielding same results – mining using linear search and mining using binary search.

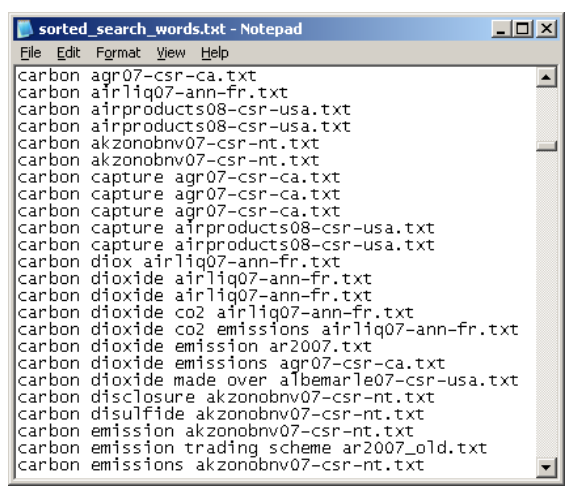

*Fig 1. Sorted keyword file*

In the first approach, the report files were read sequentially line after line and the keywords that formed the 10 categories were searched. This is good when the line numbers within a file and the total file numbers are not significantly high. The frequency count of the keywords in the 10 categories was calculated for each report file. These values were later scaled to a score of maximum 10 points per category as per table 2. In the second approach, the keywords from the report files are stored into a file, eliminating prepositions, pronouns, unwanted symbols, common verbs etc. They are written into a file in the format: <keywords, file name> as in figure 1, so that it would help in frequency calculation later. The file is later sorted based on initial word using quick sort algorithm. The keyword search is performed using binary search, which is faster than sequential search. This approach is good when the line numbers within a file and the total file numbers are very high. It does take a while to write the details into the file initially, but once that is done, the search and further calculations are very fast. The overall flow chart is shown in figure 2.

After performing the data mining technique, the reports that are in foreign languages with zero frequency (in all the variables) are eliminated and the final sample for further analysis is minimized to 539. Exploratory and confirmatory analyses were performed on the said data.

Multivariate Discriminant Analysis (MDA) is a reversal of multivariate analysis of variance (MANOVA). In MANOVA, the independent variables are the groups and the dependent variables are the groups. Where as in MDA, the independent variables are the predictors and the dependent variables are the groups and it is used to predict the membership in naturally occurring groups.

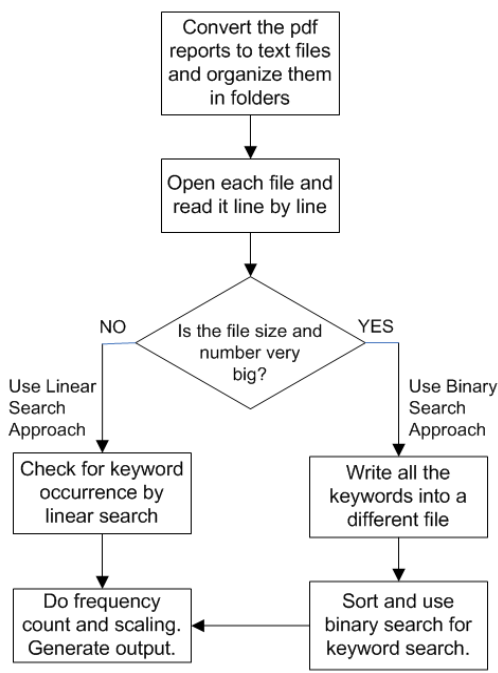

*Fig 2. Flow chart of the data mining process*

Structural Equation Modeling (SEM) estimates a series of separate but interdependent multiple regression equations simultaneously. The researcher draws upon the theory and the research objectives to determine which independent variable will predict which dependent variable. The proposed relationships are then translated into a series of structural equations for each dependent variable. The structural model expresses these relationships among independent and dependent variables. In SEM there is a unique feature of being able to include variables that are not directly measurable and are thus called unobserved or latent constructs.

## 6. Results and Discussion

The overall sample size of this study after elimination of foreign language reports is 539. The frequencies from the figure 3 shows 55.47% of reports are from secondary industry/manufacturing sector, 22.82% of tertiary services and 21.71% of primary industry respectively. The major contributors being manufacturing sector, thus confirms the findings of Kolk et al. [24], and Grey et al [25].

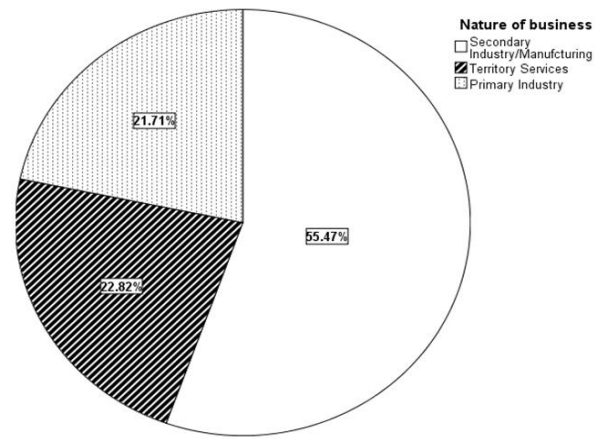

*Fig 3. Industry-wise sample*

One-way ANOVA was performed to determine that all the selected variables under study will be of equal importance in various types of industrial sectors. Results as in table 3 indicate that all the variables are significant excluding shareholder value. It signifies that majority of the variables under study have different importance among the various industrial groups. Grand mean for CEO statements on environmental issues and/or sustainable issues prove to be the effective in territory and secondary sector. Whereas, Carbon dioxide emissions, is the most concern issue in primary sector.

*Multivariate Discriminant Analysis*: Wilks Lambda test is significant for both the variables. As the Wilk's Lambda in table 4a is smaller, the more important is the independent variable to the discriminant function. Box's M Test as in table 4b, is sensitive in meeting the assumption of multivariate normality. The result indicates that groups do differ in covariance matrices, violating the assumption of Multivariate Discriminant Analysis (MDA).

Table 3: Grand Mean and One-way ANOVA results

|     | Primary Sector | Secondary Sector | Tertiary Sector | Grand Mean | F | Sig |
|-----|------|------|------|------|--------|--------|
| V1  | 4.17 | 3.48 | 4.39 | 3.96 | 11.190 | .000 * |
| V2  | 3.72 | 4.03 | 4.7  | 4.04 | 4.642  | .010*  |
| V3  | 4.13 | 4.52 | 5.08 | 4.48 | 4.755  | .009*  |
| V4  | 4.94 | 4.37 | 3.7  | 4.46 | 13.442 | .000*  |
| V5  | 0.59 | 0.26 | 0.41 | 0.43 | 13.320 | .000*  |
| V6  | 0.48 | 0.63 | 0.71 | 0.59 | 4.159  | .016*  |
| V7  | 1.68 | 2.23 | 2.27 | 2.01 | 8.533  | .000*  |
| V8  | 1.4  | 1.04 | 1.33 | 1.25 | 3.668  | .026*  |
| V9  | 0.59 | 0.63 | 0.58 | 0.6  | .155   | .857 N.S |
| V10 | 4.45 | 4.45 | 2.58 | 4.04 | 26.396 | .000 * |

\* Significant at 0.05 level
N.S. Not Significant

The classification results indicate 59.7% as in table 4c, of the cases that are correctly classified. All the three cases (industries) are equally evaluated with similar scoring, thereby proves a satisfactory discriminant analysis. This also could be observed in the canonical discriminant diagrams for selected industrial sectors as in figures 4 to 6.

Table 4: Box's M Test
(4a): Wilks' Lambda

| Test of Function(s) | Wilks' Lambda | Chi-square | df | Sig. |
|---|---|---|---|---|
| 1 through 2 | .677 | 207.176 | 20 | .000 |
| 2 | .854 | 84.127 | 9 | .000 |

(4b): Test Results

| Box's M |        | 254.359 |
|---------|--------|---------|
|         | Approx.| 2.246   |
| F       | df1    | 110     |
|         | df2    | 449035.606 |
|         | Sig.   | .000    |

(4c): Classification Results (a)

|          |       | Nature of business | Predicted Group Membership | | | Total |
|----------|-------|--------|------|------|------|------|
|          |       |        | Sec. sector | Tertiary sector | Primary sector |   |
| Original | Count | Sec.   | 129  | 55   | 41   | 225  |
|          |       | Tertiary | 51 | 114  | 32   | 197  |
|          |       | Primary | 16  | 22   | 79   | 117  |
|          | %     | Sec.   | 57.3 | 24.4 | 18.2 | 100.0 |
|          |       | Tertiary | 25.9 | 57.9 | 16.2 | 100.0 |
|          |       | Primary | 13.7 | 18.8 | 67.5 | 100.0 |

a. 59.7% of original grouped cases correctly classified.

*Structural Equation Modeling:* Subsequent to the exploring of unobservable factors, Structural Equation Modeling (SEM) was performed to the proposed relationships and they were translated into a series of structural equations for each dependent variable. The structural model expresses these relationships among independent and dependent variables. In this study, the model proposed in the framework is considered and evaluated. AMOS Graphics for Windows Version 6.0 was used to estimate the regression weight of each link (arrow) and the associated significance.

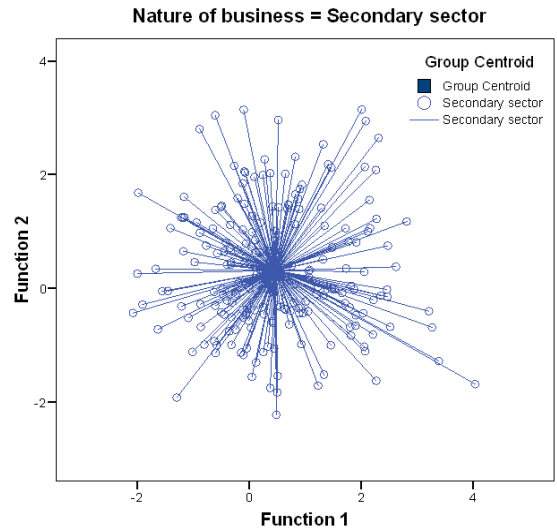

*Fig 4. Canonical Discriminant Function for secondary sector*

The convergence of the model was given by the Chi-square value, the degrees of freedom with the associated probability level and the p-value. The model was considered acceptable at the 5% level of significance if p-value > 0.05.

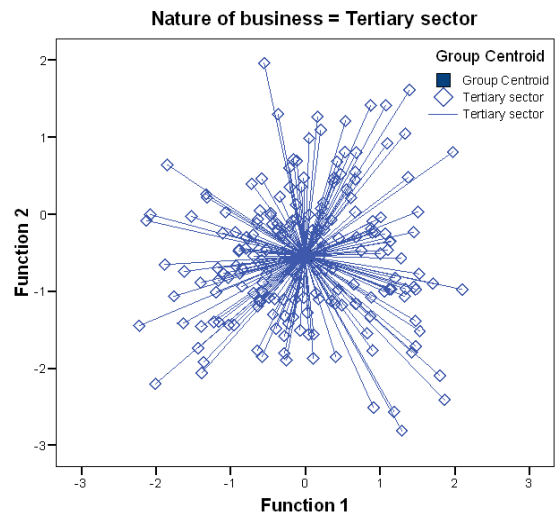

*Fig 5. Canonical Discriminant Function for tertiary sector*

In the structural equation model, as given in figure 7, the overall fit for the model is acceptable and significant as seen in chi-square/degrees of freedom = 491.659 (39, df) is 0.000 which is less than 0.05.

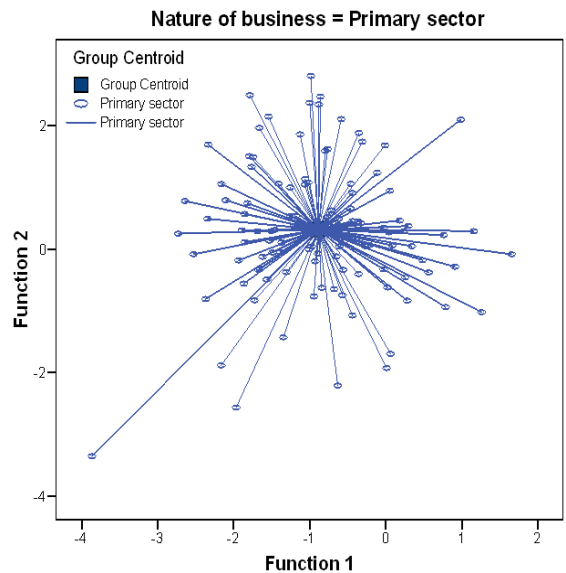

*Fig 6. Canonical Discriminant Function for primary sector*

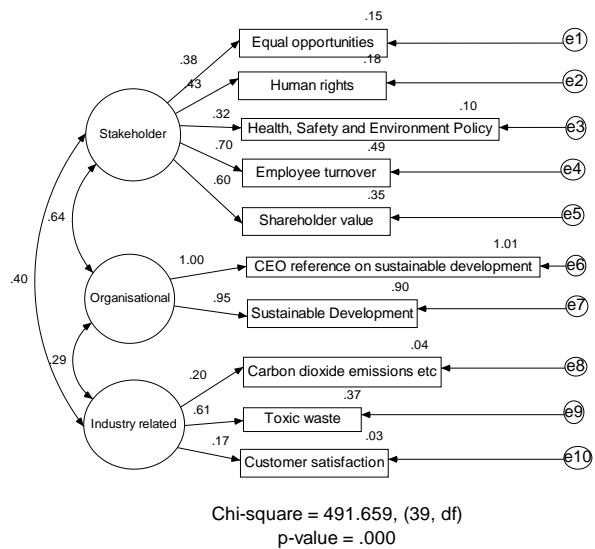

*Fig 7. Structural Equation Model – Standardized Estimates*

## 7. Limitations and Conclusion

The data mining on the qualitative reports done can be improved by making the search more intelligent when complex searches of words with similar meanings can be present within a file. Only those available with the Corporate Register website were used for this study and no additional sources were used to gather data. The dependent variable chosen for the study is type of business only. The criteria used for searching the reports, in the present study, is limited to 10 variables, this could be revised in the due course for further investigation. Similar studies could be undertaken on other demographics like, location, size, time-series, within the same domain and like wise.

The study investigates and analyses the nature of environmental reporting using data mining technique. Being vary in nature, it becomes difficult for the users of information to measure the reports and make appropriate decision making. The data taken for the study being a world directory of corporate reports, the information provided is highly reliable and up-to-date. The assessment technique developed by HBSP was used in this study to come up with the scores. 539 reports were used final analysis after removing the foreign language reports after performing the data mining technique. Nearly 55.7% of the reports are from secondary industry/manufacturing sector, followed by territory and primary sectors. One way ANOVA has revealed that all the variables selected for the study are significant excluding references to human rights, and shareholder value. Further investigation using MDA and SEM has revealed the existence of a significant relationship among the variables chosen for the model. Hence stakeholder factors, industrial factors and organizational strategic factors are significant and these findings confirm reliable direction in sustainability reporting evaluation using data mining technique.